# Shape Sensing of Variable Stiffness Soft Robots using Electrical Impedance Tomography

James Avery, Mark Runciman, Ara Darzi and George P. Mylonas, *Member, IEEE*

*Abstract*— Soft robotic systems offer benefits over traditional rigid systems through reduced contact trauma with soft tissues and by enabling access through tortuous paths in minimally invasive surgery. However, the inherent deformability of soft robots places both a greater onus on accurate modelling of their shape, and greater challenges in realising intraoperative shape sensing. Herein we present a proprioceptive (self-sensing) soft actuator, with an electrically conductive working fluid. Electrical impedance measurements from up to six electrodes enabled tomographic reconstructions using Electrical Impedance Tomography (EIT). A new Frequency Division Multiplexed (FDM) EIT system was developed capable of measurements of 66 dB SNR with 20 ms temporal resolution. The concept was examined in two two-degree-of-freedom designs: a hydraulic hinged actuator and a pneumatic finger actuator with hydraulic beams. Both cases demonstrated that impedance measurements could be used to infer shape changes, and EIT images reconstructed during actuation showed distinct patterns with respect to each degree of freedom (DOF). Whilst there was some mechanical hysteresis observed, the repeatability of the measurements and resultant images was high. The results show the potential of FDM-EIT as a low-cost, low profile shape sensor in soft robots.

## I. INTRODUCTION

Currently, the instruments used in minimally invasive surgery (MIS) are rigid and, in the case of endoscopes or laparoscopic instruments, are difficult to control due to their long length. Using stiff tools causes higher risk of unintentional damage and longer procedure times [1]. In the field of soft robotics, compliant materials are used to construct robotic devices that deform easily, making them highly applicable to interactions with soft tissue in MIS [2]. Single incision or incisionless procedures may offer faster recovery times and less patient trauma [3], so the next generation of surgical robotic devices will be designed to achieve access to all parts of the body from remote entry points, meaning they must be flexible and capable of navigating highly complex paths – all without causing damage to soft tissues [4]. However, soft robots are difficult to model, as both the structures and the methods by which they achieve locomotion or actuation of joints rely on deformable materials [5]. These materials exhibit non-linear responses to strain, so predicting the shape of soft devices and controlling the end effector is challenging, especially when external forces are exerted on the device. Control would be facilitated if the bending angles of a soft device could be accurately measured, however, current shape sensing methods have limitations.

Electromagnetic (EM) tracking can be used to track single points in space but the measurements are influenced by stray EM fields from other equipment. This makes shape reconstruction impossible and makes it difficult to integrate into an operating theatre with other electrical equipment [6]. Imaging techniques such as fluoroscopy require large doses of radiation and contrast agents, and either a C-arm or a monoplane X-ray system, while others such as ultrasound have low resolution and impair the flow of a surgical procedure [7]. Using optical tracking requires line of sight to the device and is therefore impractical or unsuitable in MIS applications. Soft strain gauges made of Eutectic Gallium Indium (EGaIn) have been used to sense contact, strain and shear deformation of the surface to which they are affixed. However, EGaIn is expensive, and elastomer strain gauges are either complex and costly to manufacture or too bulky to use in the restricted spaces imposed by MIS. Fiber Bragg gratings (FBG) were used to measure the strain between fibre segment ends embedded into a STIFF-FLOP manipulator for MIS, an elastomer continuum robot [8]. However, the location of a localized deflection cannot be determined, and the measurements are not sensitive in all directions, so a large number of fibre segments along the length of a device would be needed to fully sense the shape.

Fluidic actuation is used to produce bending motion and variable stiffness in many soft robotic devices, with internal pressure often regulated by valves and pressure sensors. Internal fluid pressure measurements from these sensors can provide a measure of bending angle or deformation of soft robots in given configurations and without contact. However, the relationship between bending angle and internal pressure breaks down when external forces are applied.

In our approach, we use saline as an electrically conductive and biocompatible actuation fluid in rapidly manufactured soft devices to enable EIT measurements. These measurements are then used to reconstruct an image of the actuator and sense its shape. The benefits of using a conductive actuation fluid are the low cost, low complexity and low profile. Additionally, the small electrodes and wires that are integrated into the soft devices described here have very little mechanical impact on the design, and the materials used can be MRI compatible. It has proved difficult to deliver shape information to clinical staff in the operating theatre using current methods, but this technique has the potential to visualize the shape of a device in real-time, which would be of use to the operators.

* Research is supported by the NIHR Imperial Biomedical Research Council (BRC), Human-centred Automation, Robotics and Monitoring in Surgery (HARMS) Lab, Dept. of Surgery and Cancer, Imperial College London, London, W2 1PF, UK. Email: james.avery@imperial.ac.uk

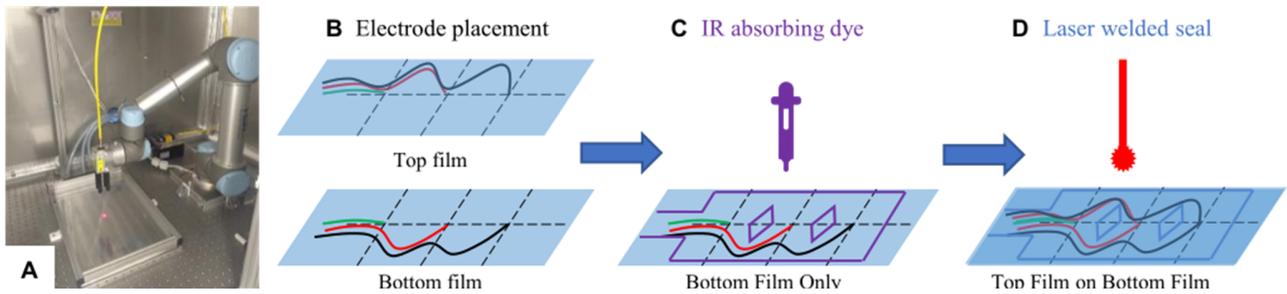

Figure 1. Manufacturing process for single layered fluidic actuated double-hinge with integrated electrodes. (A) Laser welding system, showing pilot laser (red) focused on the work area (B) Placement of electrodes (green, red and black wires) on two films to be welded. (C) Deposition of IR absorbing dye (purple) on weld pattern on bottom film. (D) Welding together top and bottom film along weld pattern (blue) using IR laser guided by the robotic arm.

EIT recordings require the application of electric currents and measurement of the resultant voltages through surface electrodes. Traditionally these measurements are used to reconstruct the conductivity distribution for the interior of the domain, with the information regarding the boundary shape typically ignored. However, there has been recent interest in reconstructing the boundary shape to improve EIT reconstructions. In the case of isotropic homogenous conductivity distribution, the electrode position, and thus the shape of the boundary can be reconstructed with an error of 5% or less [9]. There has also been considerable interest in the design of soft tactile sensors using EIT, to overcome the complications of increased internal wiring when using modular sensors [10], [11].

In this project we will investigate the use of EIT shape sensing as a potential tool for deployable soft robotics applications. Proprioceptive soft actuators with conductive fluids and single channel electrical impedance approaches have been proposed [12], [13]. However, a single measurement provides only a single degree of freedom, and limited accuracy. This project will extend the electrode count to enable full EIT methods for the first time.

## II. MATERIALS AND METHODS

### A. Robotically Guided Laser Welding

Two actuator designs that use different actuation approaches but that both contain integrated EIT shape sensors are described here. A robotically guided laser welding system developed in-house capable of welding thermoplastic films together in specific seal patterns was used to produce both devices. Depending on the seal pattern created and the construction, large volume change and bending motion can be achieved upon pressurization. In this way, the soft robotic fluidic actuators were manufactured. The planar weld patterns are generated very quickly and easily using CAD software, making this a highly adaptable, rapid and economical manufacturing method.

An infrared laser guided by a 6 DOF UR5 robotic arm (Universal Robots, Denmark) is collimated and focused with a spot size of 1 mm diameter at the interface between two triple-laminate polyethylene (PE), polyethylene terephthalate (PET), PE films within a planar work area as in Fig 1A. Thin wires are aligned with the selected weld pattern and affixed to the surface of each film, connecting to electrodes at specific locations as in Fig 1B. In this case, we placed the electrodes at points on the weld pattern that would produce the largest deformations in the finished soft actuator when pressurized. Multiple electrodes can be placed in a single chamber, which permits improved shape reconstruction of the final design using EIT imaging techniques.

In the next step, an infrared (IR) absorbing dye was deposited by using the robotic arm to guide an adapted pen where the two plastic films were to be welded together, as in Fig 1C. The robotic arm then guides the focused IR laser beam along the weld path, as in Fig 1D, which is absorbed by the dye causing local heating and bonding as the films are clamped together. The excess plastic film is then simply cut from around the welded device and the fluid inlet tubes connected. Feature sizes down to 1 mm are possible due to the spot size of the laser, precision of the robot and melting of the materials.

Multiple chambers can be welded into a single layer, and multiple layers can be stacked and welded together after an alignment step. A variety of designs for fluidic actuators with integrated EIT sensors can be produced using this approach, thereby enabling an enormous range of applications. Two soft actuators, Fig 2, were manufactured to illustrate the flexible manufacturing process: a double-hinged design on a single layer and single chamber, and a finger-like design with two layers and three chambers across those layers.

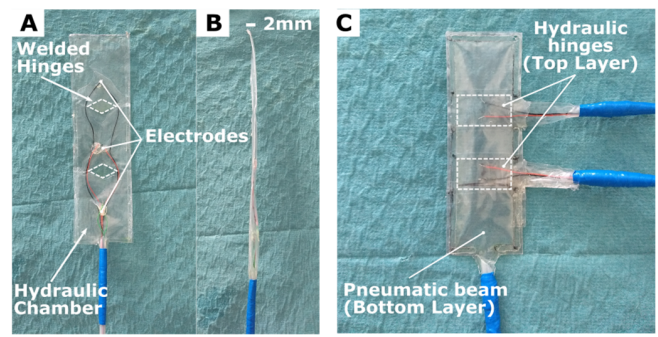

Figure 2. Double hinged actuator and double hinged pneumatic finger. (A) Double hinged actuator showing diamond shaped hinges and electrodes (green, red and black wires). (B) Side view showing low profile. (C) Pneumatic chamber and hydraulic hinges of finger design.

*1) Hinged Actuator*

The first actuator we produced is based on the fluid actuated hinges described in Ou et al. [14] that bend to an angle pre-defined by the aspect ratio of the diamond weld pattern at their centre. In our design, a chain of two hinges were connected in series to produce a curling actuator that can be seen in Fig 2A. The double-hinge actuator measures 200 mm x 50 mm and has a very low profile, as shown in Fig 2B. The two diamond shaped hinges measure 20 mm x 10 mm. Three electrodes were placed on each of the top and bottom films used to make the actuator, at the points that would give the highest deformation after pressurization, as in Fig 1B. After the laser welding stages, the design was cut out and filled with 0.1% saline solution that would act as both actuation fluid and conductive medium for the EIT recordings. A syringe pump is connected and used to pressurize the hydraulic chamber.

*2) Finger Actuator*

Another actuator was fabricated, which used the method detailed by Niyama et al. [15] to bend a pneumatic finger. The finger, in Fig 2C, consists of a pneumatic beam formed by an 85 mm x 25 mm rectangular chamber welded on one layer, with two 10 mm x 25 mm hydraulically actuated chambers spaced 25 mm apart and 30 mm from each end on another layer, which was aligned with and welded onto the first. When the pneumatic finger is pressurized below 0.5 bar in order to remain compliant, the hydraulic chambers cause buckling and bending of the pneumatic beam due to the deformation introduced when they are pressurized. Two electrodes were positioned in the area of the largest expected change in cross-sectional area, across the centre of the chamber. As with the first actuator, the hydraulic chambers were filled with 0.1% saline solution and a syringe pump connected.

*B. Electrical Impedance Tomography – Hardware*

An EIT system comprises three major components, an alternating current source, a voltage recorder and electrode array with a method of addressing specific electrodes for current injection. A new EIT sensor, Fig 3, based on the UCL ScouseTom EIT device [16], was designed with a focus on data acquisition speed, and compatibility with robotic control systems. Typical EIT systems inject current between sequential pairs of electrodes, known as Time Division Multiplexing (TDM). However, the need to switch between electrode pairs imposes a limit on the data acquisition rate. This is exacerbated by the capacitive electrode-electrolyte interface, which introduces a "settling time" after switching current injection pairs. Common techniques used to minimize this effect, such as the use of large Ag/AgCl Electroencephalogram (EEG) style electrodes, may not always be feasible in a soft robotics context.

By injecting multiple frequencies simultaneously through different electrode pairs, Frequency Division Multiplexing (FDM) removes the need to switch entirely, making it a suitable technique for shape sensing [17]. This was implemented using a parallel system with six independent current sources, using the differential Howland current pump configuration (Appendix). Each source can inject current at 165 µA between 10 Hz to 100 kHz between a single pair of electrodes, well below the IEC 60601 safety limits for medical electrical equipment [18]. Voltages were recorded using a National Instruments USB-6216 16-bit DAQ at 50 kHz sampling rate. These were then connected to the electrodes using a custom PCB. To obtain a single EIT measurement the voltage recorded on a single electrode is demodulated at a single frequency and averaged over several sine wave periods. The range of 3 ms to 250 ms could be selected based on given limitations of USB data transfer speeds and DAQ buffer size.

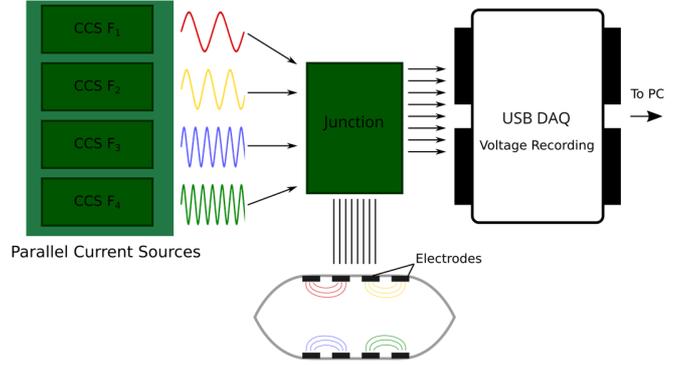

Figure 3. Overview of Frequency Division Multiplexed EIT system, for simultaneous current injection at different frequencies. Parallel constant current sources CCS and USB DAQ voltage recorder.

*C. Electrical Impedance Tomography – Forward and Inverse Solution*

The forward problem in EIT is a combination of the quasi-static Maxwell equations with the addition of the Complete Electrode Model boundary conditions for a single pair of electrodes (l = 1,2) [19]

$$\nabla \cdot \sigma \nabla u = 0, \quad (1)$$

$$\sigma \frac{\delta u}{\delta n} = 0 \quad \text{off} \bigcup_{l=1}^{2} e_l \quad (2)$$

$$u + z_l \sigma \frac{\delta u}{\delta n} = V_l, \quad l = 1,2 \quad (3)$$

where $\sigma$ is the conductivity, $u$ is the electric potential, $n$ is the normal to the boundary, $e_l$ is the $l$th electrode boundary, $z_l$ is the contact impedance on the $l$th electrode, and $V_l$ are the measured voltages. The finite element method is then employed to solve a discretized form of (1) for a specific domain. This is then repeated for all injection pairs and measurement electrodes. Typically, a linear relationship between voltage changes and conductivity changes is made, allowing for a construction of a Jacobian matrix $J$, which maps the changes in conductivity in each element to changes in voltage on all electrodes:

$$\delta u = J \delta \sigma \quad (4)$$

However, as the aim is to reconstruct conductivity changes from voltages measurements, (2) must be inverted. However, as $J$ is ill-conditioned, regularization is necessary before it can be inverted. Zeroth order Tikhonov

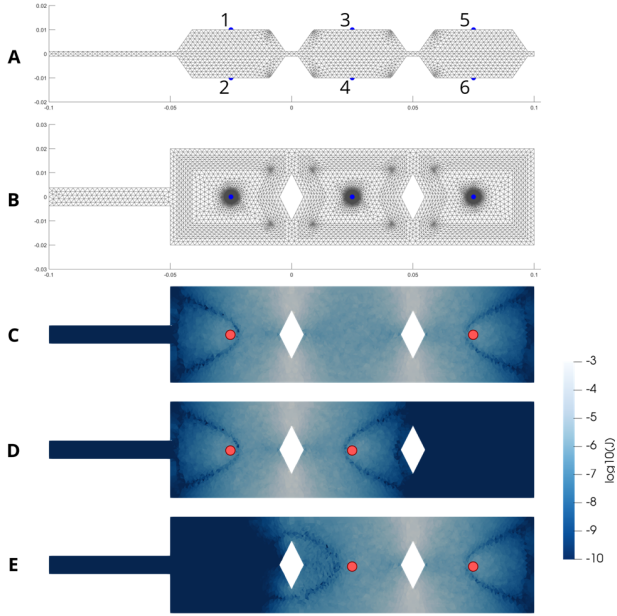

Figure 5. EIT forward model for double hinged actuator (A) FEM with six electrodes, (B) Refinement around electrodes and diamond weld pattern and (C-E) sensitivity $J$ matricies for three injection pairs (red dots)

regularization was employed which minimizes the magnitude of the conductivity changes in the mesh:

$$\widehat{\delta\sigma} = \arg\min_{\delta\sigma} \|\delta v - J\delta\sigma\|^2 + \lambda\|\delta\sigma\|^2 \quad (5)$$

$$J_\lambda^{-1} = (J^T J + \lambda I)^{-1} J^T \quad (6)$$

Where $\widehat{\delta\sigma}$ is an estimated conductivity change, $\delta v$ are the boundary voltage change and $\lambda$ is a hyperparameter controlling the extent of the regularization. The reconstructions were performed using the method developed by Aristovich et al. [20], with $\lambda$ selected through cross-validation in a hexahedral subdomain (Appendix).

Solid model approximations of the inflated actuators were designed in SolidWorks and used to create tetrahedral FEMs in COMSOL (Comsol Ltd., UK), Fig 4A. The mesh was further refined around each electrode, as in Fig 4B. To visualize the sensitivity of each measurement, the matrix $J$ was calculated for each injection pair and corresponding measurement, as shown in Fig 4C-E.

### D. Measurement Setup

In the Double Hinge Actuator, current was injected between chambers, to maximize the sampling volume. Given the symmetry of the design, Fig 4A, this gave three possible injections pairs. Current was injected at 2, 4 and 6 kHz, between the pairs of electrodes shown in Fig 4C-E. To maximise the sensitive volume, current was injected between electrodes on opposite surfaces, i.e. 1-6, 2-3 and 4-5. Voltages were recorded between the electrodes opposite to the recording sites: 2-5 1-4 and 3-6. This resulted in three voltages for each injection for a total of nine measurements.

As the chambers in the finger actuator are independent, only a single two channel impedance measurement was possible for each. The separation of the two injection frequencies was increased to 10 kHz to minimize capacitive coupling.

### III. EXPERIMENTAL SETUP

#### A. EIT System validation

To test the frequency invariance of the EIT device, measurements were made on a resistor phantom (SwissTom Ag, Switzerland), six equally spaced injection frequencies between 2 kHz and 12 kHz.

#### B. Hinged actuator

The sensitivity maps in Fig 4C-E clearly show the sensitivity is focused at the narrow points between the diamond welds, where the hinge axis is located (Fig 2A). Further the injections between neighbouring chambers, Fig 4D and E, are largely independent of each other as there is no sensitivity to changes in the other chamber. Injecting current across the whole actuator (Fig 4C), ensures there is information regarding the ratio of the two hinge axes in each recording.

Voltage changes during manual rotation of each axis in turn, Fig 5, correlates with the sensitivity maps. There are large changes observed in the corresponding channels when the first hinge is moved, but negligible changes when the other hinge is rotated. However, for the injection across the whole sensor, changes are seen in every case.

To test the repeatability of the measurements and resulting reconstructions, each hinge was bent through 0-90° in 10° intervals, with the other held straight, Fig 6. Each recording was repeated three times, with the actuator fully pressurised. Finally, recordings were taken during repeated pressurization with the actuator allowed to move freely, Fig 7.

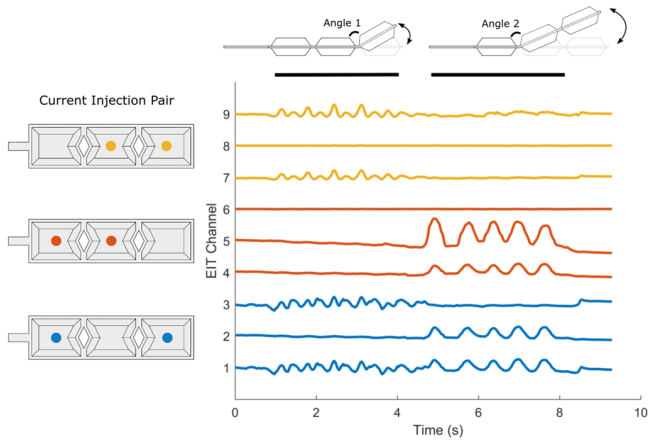

Figure 4. EIT measurements during isolated sequential rotation of each DOF. Rotation was performed by hand, and repeated five times. Three voltages are recorded for each current injection. Injections across chambers are largely insensitive to rotation in the other DOF.

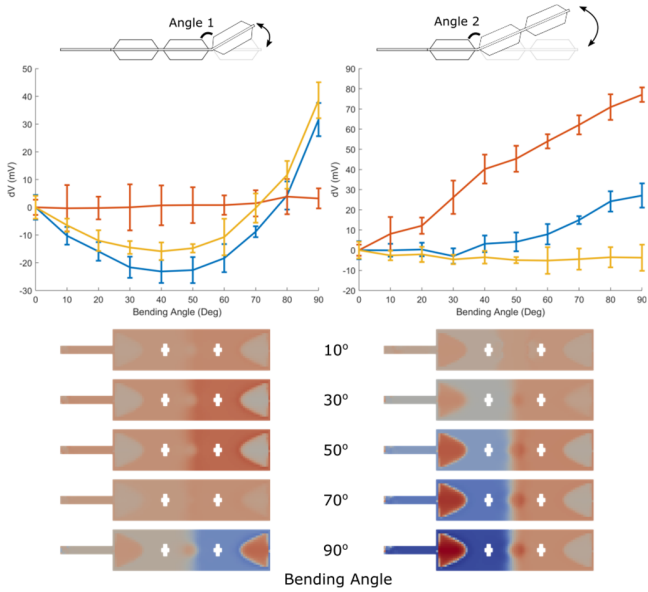

Figure 7. Change in voltages with bending angle. (Top) Mean±SD change for each injection pair in Fig 5, for both hinges, (Bottom) EIT reconstructions with changing bending angle

*C. Double-joint finger bending sensor*

The sensitivity map of a single chamber, Fig 8b, shows that whilst the measurement is sampling near the whole chamber, it is focused at the centre, at the site of the greatest cross-sectional change. Data were collected for pressurization of each chamber individually, and then for both chambers simultaneously.

## IV. RESULTS

*A. EIT system*

The shortest possible acquisition time was 3 ms, however, the temporal resolution was limited to 20 ms due to USB data transfer speeds. The average SNR in the resistor phantom measurements was 65.2 ± 0.8 dB. There were no significant differences between data collected at all frequencies (P < 0.0338). The saline chambers in the actuators presented a greater range of loads, 2.3 kΩ to 4.2 kΩ compared to 171 Ω to 476 Ω in the resistor phantom. However, the SNR was not adversely affected with a mean of 66.0 ± 7.6 dB across all recordings.

*B. Hinged actuator*

Deflection of each hinge axis resulted in a distinct profile against angle, Fig 6. There was an approximately linear increase in the two sensitive injections for Angle 2, $R^2$ =0.81 and 0.99 respectively. Changes in Angle 1 displayed an initial decrease, before sharply increasing above 50° as the cross section of the hinge decreased. These two distinct modes are shown in the EIT reconstructions. Changes to Angle 2 gave a large decrease in intensity in the closest chamber, with the far chamber largely unchanged. Deflection of Angle 1 give rise to increased intensity in the respective chamber for small angle changes, before changing to decreased intensity at extreme deflections. The amplitude of the voltage changes were also not consistent between changes in Angle 1 & 2, with 38.8 mV and 88.1 mV respectively. This is likely a result of the increased wiring within the hinge at Angle 2, so the geometric changes are a greater proportion of the cross-sectional area which produces greater voltages.

The voltages recorded during actuation show a more complicated response than the individual DOF in Fig 7. Changes in Angle 2 dominate, as the majority of the voltages decrease with increasing pressure. This is likely a result of greater changes in the chamber furthest from the inlet, which appeared to change cross section the most during actuation. The EIT reconstructions were reproducible across repeated actuations but cannot readily be understood as linear combinations of the images of the separate bending modes in Fig 6.

*C. Finger actuator*

The signals recorded in each chamber were decoupled, Fig 8C, with changes only observed in the actuated chamber. The signals were similarly independent during dual pressurization. The EIT reconstructions were repeatable, and unlike the hinged actuator, the images were reproducible when both chambers were active. However, the low electrode count does not produce images beyond mere reproductions of the sensitivity map shown in Fig 8B.

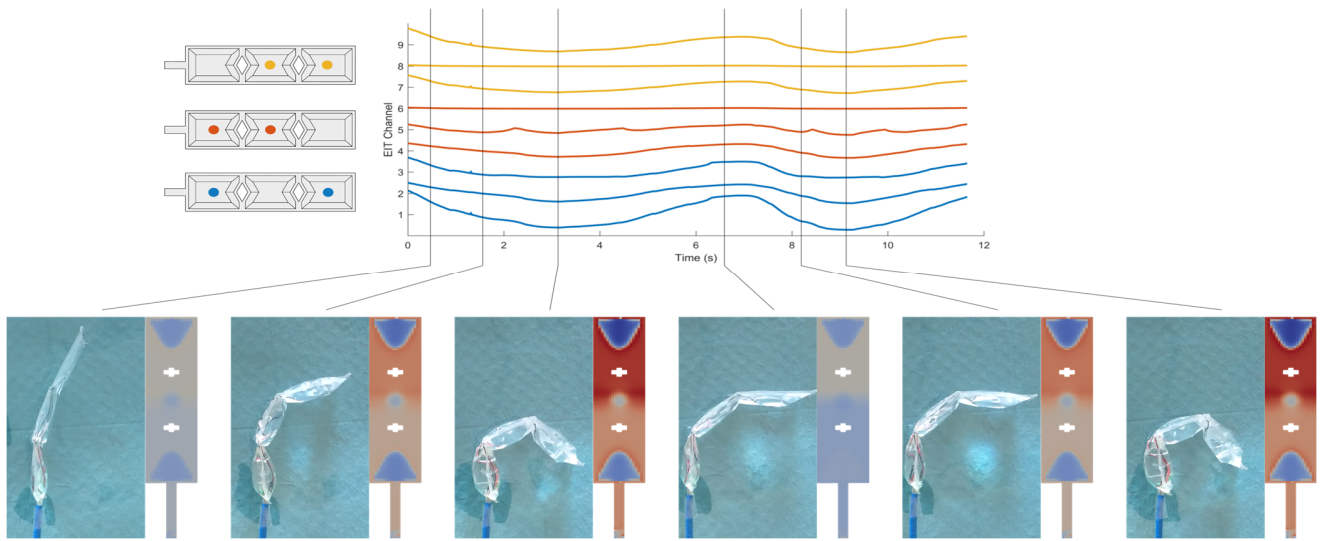

Figure 6. Changes in EIT measurements during pressurisation of double hinge actuator, with corresponding images and EIT reconstructions

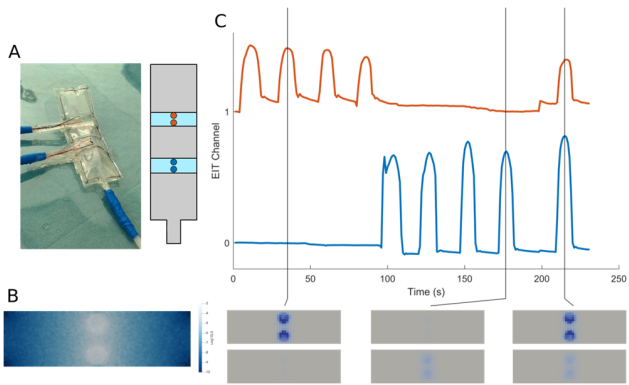

Figure 8. EIT reconstructions in finger actuator (A) Finger actuator with electrodes in hydraulic chambers (B) Sensitivity map *J* and (C) Voltages recorded during actuation of each chamber sequentially, and finally both chambers simultaneously

### D. Mechanical Performance

Upon pressurization, both of the hinges of the double-hinged actuator increased their bending angle at an approximately equal rate. When the pressure was decreased, significant hysteresis was observed as the bending angle of each hinge did not return to zero. This is evident in Fig 6. Hysteresis was not observed on the finger-like actuator because the pneumatic beam at the centre of the design tended to stay straight but the maximum bending angles achieved by each actuator finger-like actuator were limited to less than 30°.

## V. DISCUSSION

The results demonstrate FDM-EIT gives a reproducible response to shape changes in soft robotic actuators. The imaging results contain more DOF than single channel electrical impedance (EI) measurements, comparable to multi-channel FBG sensors, with a lower profile and less impact on mechanical performance. Unlike the individual DOF images, the more complex shape changes, Fig 7, were not easily identifiable visually. This suggests a further feature extraction and classification step would be beneficial, as well as higher electrode counts.

The results of the sensitivity analysis, Fig 4A, are highly dependent upon the accuracy of the model used. In this study simple approximations were used, resulting in artefactual changes in EIT reconstructions, particularly in the Hinged Actuator, Fig 6 and 7. These could negatively impact any subsequent classification steps based on these images and would be reduced using representative models. However, given the complexity of the deformation of these actuators, either a dedicated deformation model [14] or 3D surface scans would be required.

Currently, the FDM-EIT system has a temporal resolution of 20 ms, and EIT images can be reconstructed in ~50 ms. Therefore, either voltage or image outputs have the potential for closed loop control of shape for soft robots. Using AC currents allows for greater injection amplitudes than DC under IEC60601 regulations, which benefits SNR. This also minimizes electrode polarization problems experienced in other EI sensors [21]. The FDM-EIT approach has benefits in an MIS context as it uses safe, insensible currents, and the system could be made MRI compatible in the same manner as MRI-EEG systems through the use of "MR-safe" non-magnetic electrode materials.

Using saline solution as the actuation fluid helps to maintain a low-profile design, however, it adds weight in comparison with pneumatic actuation. The power density is higher than pneumatic actuation although the response times are generally longer. Opting for a less voluminous design would reduce weight without impacting on the capabilities of the EIT sensor. When pressurized with the actuator configured to bend upwards against gravity, Angle 2 of the double-hinged actuator increased but Angle 1 failed to lift itself, suggesting a higher actuation pressure is necessary to fully actuate. This imposes higher requirements on the burst pressure achieved using laser welding, which would also be beneficial to ensure safety.

## VI. CONCLUSION AND FUTURE WORK

The principle of a shape sensor for soft robots using FDM-EIT has been demonstrated in two soft actuators. The sensors have been integrated into a versatile manufacturing process and are thus adaptable to a number of applications. Through the selection of electrode location and current injection patterns, the sensitivity of the sensor was targeted at the regions of highest deformation.

Whilst the actuators demonstrated the principle of EIT shape sensing, there is potential for improvement. Smaller, lighter designs would improve the mechanical performance, and a greater density of electrodes would improve the reconstructed images. Reducing modelling errors using more representative FEMs would also be advantageous in both the initial optimisation and subsequent imaging stages. The sensing outputs seem a good candidate for machine learning classification techniques, however there is investigation required as to whether the voltages or subsequent image data would be the most robust.

Further, the designs described here can also be used as passive sensors as opposed to actuators. If attached to an active mechanism, the shape could be determined by reading from the EIT sensors on the soft passive sensor.

## APPENDIX

EIT reconstruction software, and schematics for the parallel current source can be found at https://github.com/EIT-team